\pdfoutput=1
\documentclass{article}
\usepackage{spconf,amsmath,graphicx}
\usepackage{cite}
\usepackage{gensymb}
\usepackage{multirow}
\usepackage{amssymb}
\usepackage{xspace}
\usepackage[table]{xcolor} 

\newcommand{\empmethodname}{GaitGL-\textbf{M}\xspace}
\newcommand{\methodname}{GaitGL-M\xspace}
\newcommand{\dynamicmodule}{SiMo\xspace}
\newcommand{\enhancemodule}{FeMo\xspace}

\newcommand{\mat}[1]{\mathbf{#1}}
\newcommand{\conv}{\mathsf{Conv}}

\newcommand{\gap}{\mathsf{GAP}}
\newcommand{\seqLength}{K}
\newcommand{\clipLength}{{\lfloor K/L \rfloor}} 
\newcommand{\best}[1]{\textbf{#1}}
\newcommand{\better}[1]{\underline{#1}}
\newcommand{\cc}{{}}

\newlength\savewidth\newcommand\shline{\noalign{\global\savewidth\arrayrulewidth
  \global\arrayrulewidth 1.5pt}\hline\noalign{\global\arrayrulewidth\savewidth}}
  
\newlength\lighterwidth\newcommand\lhline{\noalign{\global\savewidth\arrayrulewidth
  \global\arrayrulewidth 1.0pt}\hline\noalign{\global\arrayrulewidth\savewidth}}

\title{
Motion Matters: A Novel Motion Modeling For Cross-View Gait Feature Learning}

\name{Jingqi Li$^{\star}$ \qquad Jiaqi Gao$^{\star}$ \qquad Yuzhen Zhang$^{\star}$ \qquad Hongming Shan$^{\sharp}$ \qquad Junping Zhang$^{\star\dagger}$\thanks{$^{\dagger}$: Corresponding author} }

\address{$^{\star}$ Shanghai Key Lab of Intelligent Information Processing, School of Computer Science\\
$^{\sharp}$ Institute of Science and Technology for Brain-inspired Intelligence\\
Fudan University, Shanghai 200433, China
}

\begin{document}

\maketitle
\begin{abstract}
As a unique biometric that can be perceived at a distance, gait has broad applications in person authentication, social security and so on. Existing gait recognition methods suffer from changes in viewpoint and clothing and 
barely consider extracting diverse motion features, a fundamental characteristic in gaits, from gait sequences. This paper proposes a novel motion modeling method to extract the discriminative and robust representation. Specifically, we first extract the motion features from the encoded motion sequences in the shallow layer. Then we continuously enhance the motion feature in deep layers. This motion modeling approach is independent of mainstream work in building network architectures. As a result, one can apply this motion modeling method to any backbone to improve gait recognition performance. In this paper, we combine motion modeling with one commonly used backbone~(GaitGL) as \methodname to illustrate motion modeling. Extensive experimental results on two commonly-used cross-view gait datasets demonstrate the superior performance of \methodname over existing state-of-the-art methods.

\begin{keywords}motion modeling, plug-and-play
\end{keywords}

\end{abstract}
\section{Introduction}
\begin{figure*}[ht]
   \centering  \includegraphics[width=\linewidth, height=0.25\linewidth]{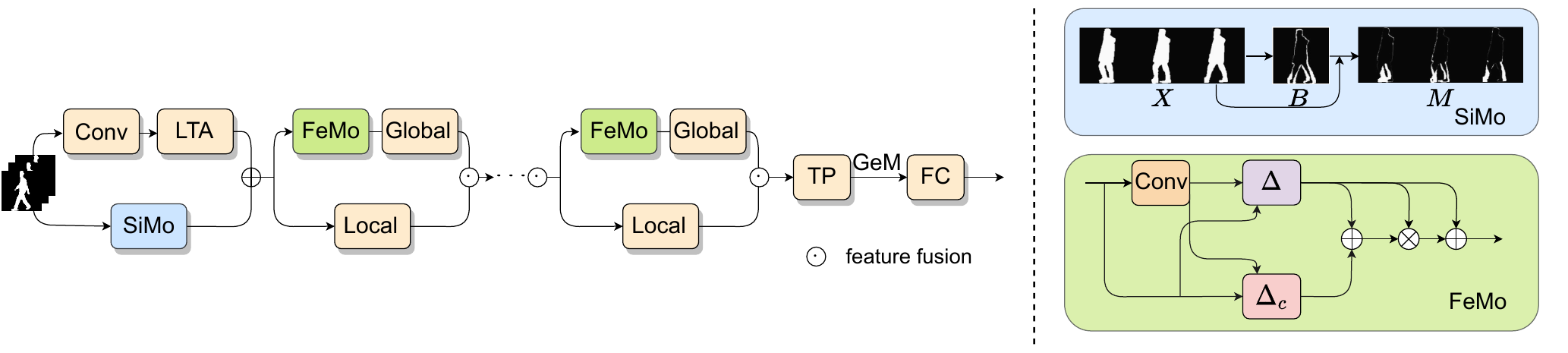}
   \caption{The picture on the left is the overall framework of \methodname. Removing \dynamicmodule and \enhancemodule for motion modeling leaves the architecture of GaitGL~\cite{gl_LinZ021_iccv}. The upper part of the right image is the motion sequence construction process in \dynamicmodule, where $X$ is the silhouette sequence, $B$ is the mask, and $M$ represents the generated motion sequence. The lower part is the forward path in \enhancemodule, and the backward direction is treated similarly.}
   \label{figures: model}
\end{figure*}

Silhouette, a standard modality for appearance-based gait recognition, is a binary map generated by segmenting the individual and background. However, the silhouettes among \textit{different individuals} only have subtle variances when the body shapes are similar, inducing the nondiscriminative of the appearance-dependent gait feature. On the contrary, the walking speed and gait cycle are distinguished even though the body shapes look similar among these individuals. Additionally, the silhouettes of \textit{one individual} visually differ when the clothing or viewpoint varies, revealing the vulnerability of the appearance-dependent gait feature. Nevertheless, this person's motion information, such as speed and gait cycle, remains consistent. Fortunately, this motion information is reflected in the frame-to-frame changes in the sequence of silhouettes, which can be explored to obtain discriminative and robust gait features.

Recent works mainly aggregate the sequences in different stages. Template-based methods~\cite{gei_HanB06,cgi_WangZPYW10_eccv,geni_BashirXG09,feni_ChenLZHT09} compress all silhouettes into one gait template before extracting features, sacrificing the essential temporal information. Set-based methods~\cite{gaitset_ChaoHZF19, gln_hou2020_eccv} rather aggregate after the feature extraction stage by pooling. More recently, many new works~\cite{gaitpart_FanPC0HCHLH20_cvpr, mt3d-LinZB20-mm,3dlocal_huang2021_iccv, cstl_HuangZWWYH0021_iccv,gl_LinZ021_iccv} further aggregate the feature sequence in the feature extraction stage using temporal convolution. However,  it is hard to  extract the motion information through temporal aggregation. As one recent work claimed~\cite{lagana_cvpr22_chai}, only relying on temporal convolution is not enough to ensure the uniqueness of the extracted gait feature, let alone the temporal pooling. But, it is noticeable that  its theoretical analysis proves that the relationship between adjacent frames can provide the distinguishability of features.

Motivated by these observations, we propose a novel motion modeling for gait recognition, through utilizing the motion information inherent in silhouette sequence and enhancing the motion information in gait representation. Unlike the prior work that employs local self-similarities as the motion information~\cite{lagana_cvpr22_chai}, we define the motion information as the holistic temporal changes of all body parts. Our motion modeling method mainly comprises a \textbf{Si}lhouette-level \textbf{Mo}tion extractor~(\textbf{SiMo}), which facilitates silhouette motion encoding, and a \textbf{Fe}ature-level \textbf{Mo}tion enhancement~(\textbf{FeMo}), which preserves feature-level motion details. 
This motion modeling method is applicable to any existing backbone. To better illustrate its usage, we plug the SiMo and FeMo into GaitGL~\cite{gl_LinZ021_iccv}, named \empmethodname. Additionally, the performance of plugging these two modules into GaitSet~\cite{gaitset_ChaoHZF19} is presented in experiments~(see Table~\ref{table: network_ablation}).

The contributions of this paper are summarized as follows. 1) We propose a novel motion modeling method to extract the discriminative and robust gait representation. Moreover, this method is independent of network architecture. Thus one can plug it into any existing backbone.
2) We propose two plug-and-play modules in motion modeling, including a silhouette-level motion extractor and feature-level motion enhancement. Additionally, we combine them with one of the most popular backbones in gait recognition GaitGL as \methodname to show the effectiveness of our proposed modules. 
3) Extensive experimental results demonstrate the proposed \methodname's superiority in the CASIA-B and OU-MVLP datasets, especially when spatial variations appear.

\section{Methodology}

Figure~\ref{figures: model} presents the framework of \methodname, a combination of our motion modeling and one of the most popular backbones-GaitGL. In \methodname, the silhouette-level motion extractor is paralleled with the original feature extractor, and the features extracted by these two branches are concatenated together as the subsequent input. The feature-level motion enhancement is used to preserve motion details before feature extraction in the global branch. 

Before detailing the key modules in \methodname, we provide the description of the gait data. We denote a gait sequence of $ \seqLength $ frames as $\mat{X} = [\mat{S}_1,\mat{S}_2, \ldots, \mat{S}_\seqLength] \in \mathbb{R}^{\seqLength \times C \times H \times W}$, where $C$, $H$, and $W$ represent the number of color channels, height, and width of one frame $\mat{S}$, respectively. Instead of using RGB video frames, the most popular modality used in gait recognition is the silhouette, which is a binary image highlighting the region of the person; correspondingly, $C=1$ for silhouettes.

\subsection{Silhouette-level Motion Extractor~(SiMo)}

The \dynamicmodule explicitly ensures the network's perception of motion information. It first constructs the motion sequences by seeing the dynamic region exclusively and then extracts the shallow motion feature. Finally, we concatenate the motion feature with the appearance feature. In this way, the sparse motion won't be diluted by the dense appearance feature.

{ \noindent \emph{Motion sequence construction:}}\quad
During a gait cycle, the limbs move alternately. It means that the limbs move for a period of time and are relatively static for a period of time in a gait cycle. 
Considering the intermittent nature of the limb's motion, we establish the motion sequence at temporal clips rather than the entire sequence to increase the representational space of motion. In addition, the process of generating silhouette maps contains some noise. Thus, direct performing motion filtering on the whole sequence introduces pseudo-motion information. And selecting the motion signals from temporal clips can also diminish this problem.

More specifically, we uniformly divide a given gait sequence into clips $\mat{C}\in\mathbb{R}^{\clipLength \times L \times H\times W}$ along the temporal dimension. Then, the motion region mask $\mat{B}_i$ for $i$-th clip can be generated by 
\begin{align}
    \mat{B}_{i} = \max(\mat{C}_i) - \min(\mat{C}_i),
\end{align}
where the $\max$ and $\min$ denote the maximum and minimum value of one spatial position among $L$ frames for a clip, and $\mat{B}_{i}\in\mathbb{R}^{H\times W}$.
Obviously, we can get the motion clip by multiplying each silhouette in the clip with its corresponding mask in an element-wise manner, defined as follows:
\begin{align}
    \mat{M}_i = \mat{B}_i \odot \mat{C}_i = [\mat{B}_i\mat{S}_{(i-1)L+1}, \ldots, \mat{B}_i\mat{S}_{(i-1)L+ L}].
\end{align}
Then, concatenating the motion clip along the temporal dimension yields the final motion sequence. 
The constructed motion sequence serves as the supplement input to the original silhouette sequence, which could strengthen the robustness of gait representation against the spatial variants.

{\noindent \emph{Motion feature extraction:}}\quad
In order to extract the shallow motion feature, we first aggregate the frame-level motion within a short clip to obtain $\overline{\mat{M}}_i$. 
Then, all of them are concatenated on the temporal dimension, denoted as $\overline{\mat{M}}$.
Subsequently, we feed the aggregated motion sequence $\overline{\mat{M}}$ into the convolution layer for extracting the motion feature $\mat{F}^m$ acquainted with temporal evolution.

\subsection{Feature-level Motion Enhancement~(FeMo)}

The stacked temporal convolutions aggregate the adjacent frames and may weaken the inter-frame differences, making the motion information hard to model. A natural idea to address this issue is to boost the motion-related information before each temporal convolution operation. Inspired by motion encoding in action recognition~\cite{stm_iccv19_jiang, tea_cvpr20_li}, we regard motion intensity as an attention map to recalibrate the original features and enhance motion-related features. The significant difference from action recognition is that the only action in gait is walking. We aim to mine different walking patterns to identify identities, so fine-grained motion information is necessary. 

Firstly, we introduce a bi-directional fine and coarse temporal difference module to distill the subtle motion information from the feature volume. Enlarging the motion search space allows us to understand the importance of different motion directions, and we enlarge the space by spatial convolution.

Consequently, the temporal difference is formulated as:
\begin{align}
    \Delta(\mat{G}_t, \mat{G}_{t+1}) = \conv_\textsc{2D}(\mat{G}_{t+1}) - \mat{G}_t,
\end{align}
where $\conv_\textsc{2D}$ denotes 2D convolution of size $3\times3$. Take a single channel for example, the convolution kernel element $w_{i,j}$ semantically represents the importance of different motion directions $(i,j)$. The larger the value, the higher the degree of attention to motion in the direction. 
After encoding the fine-grained motion information of each cell, we summarize the spatial information to represent the coarse motion of the whole body by global average pooling: 
\begin{align}
    \Delta_\text{C}(\mat{G}_t,\mat{G}_{t+1}) = \gap(\Delta(\mat{G}_t,\mat{G}_{t+1})),
\end{align}

Moreover, we utilize bi-directional temporal differences to enhance the richness of motion information expression. Overall, the temporal differences are formulated as follows:
\begin{align}
\begin{cases}
\mat{D}_t^\text{F}=\Delta(\mat{G}_t,\mat{G}_{t+1}) + \Delta_\text{C}(\mat{G}_t,\mat{G}_{t+1}), \\
\mat{D}_t^\text{B}=\Delta(\mat{G}_{t+1},\mat{G}_t) + \Delta_\text{C}(\mat{G}_{t+1}, \mat{G}_t). 
\end{cases}
\end{align}

Here the superscripts $\text{F}$ and $\text{B}$ denote the forward and backward operations respectively.

Secondly, we recalibrate the module guided by motion information. 
A sigmoid function $\sigma$ is utilized to map the motion intensity into range $(0,1)$, yielding the average attention from forward and backward directions: 
\begin{align}
    \mat{W} = (\sigma(\mat{D}^\text{F}) +  \sigma(\mat{D}^\text{B}))/2
\end{align}
When conducting recalibration, the input feature performs addition with the motion feature, which is the element-wise product of the input feature and motion-aware attention, followed by a convolutional layer to extract motion-aware spatiotemporal feature:

\begin{align}
\mat{G}^m &= \conv(\mat{G}+\mat{G}\odot \mat{W}). 
\end{align}

\subsection{Feature Mapping and Loss Function}
After the whole feature extraction stages, we employ temporal max pooling to aggregate the feature volume, 
followed by generalized-mean pooling~(GeM) for spatial pooling~\cite{GeM_radenovic2018fine}. Afterward, we utilize separate fully-connected layer and batch normalization layer to map the feature into a metric space. 
Following GaitGL~\cite{gl_LinZ021_iccv},  we use triplet loss and cross entropy loss function to optimize \methodname.

\section{Experiments}
\subsection{Datasets}
\textbf{CASIA-B} dataset~\cite{casia-b-yu2006} is a widely used dataset containing $124$ individuals. There are $11$ camera-perspective uniformly sampling from range $(0\degree,180\degree)$ with $10$ sequences in $3$ walking conditions for each individual. Normal status~(NM) has $6$ sequences, bag carrying~(BG) and coat-wearing~(CL) have $2$ sequences respectively. Under the subject-independent protocol~\cite{survey_pami2022_sepas}, we use the large-sample training~(LT) strategy~\cite{gaitset_pami21_chao}, in which the sequences are from $74$ different identities. \textbf{OU-MVLP} is one of the biggest cross-view dataset~\cite{oumvlp-takemura2018multi} with $10,307$ individuals. There are $14$ views sampling from $(0\degree,90\degree)$ and $(270\degree, 360\degree)$ respectively per subject and $2$ sequences (\#seq-$00$, \#seq-$01$) per view. The train data contains $5,153$ individuals, and another $5,154$ individuals are taken as test data. In the testing phase, we set \#seq-$01$ as gallery data.

\begin{table*}[h]
\centering
\resizebox{0.8\textwidth}{!}{
\begin{tabular}{cc|ccccccccccc|c}
\shline
\multicolumn{2}{c|}{Gallery NM\#1-4} & \multicolumn{11}{c|}{0\degree-180\degree} & \multirow{2}{*}{Mean}  \\ \cline{1-13} 
\multicolumn{2}{c|}{Probe} & \multicolumn{1}{c|}{0\degree} & \multicolumn{1}{c|}{18\degree} & \multicolumn{1}{c|}{36\degree} & \multicolumn{1}{c|}{54\degree} & \multicolumn{1}{c|}{72\degree} & \multicolumn{1}{c|}{90\degree} & \multicolumn{1}{c|}{108\degree} & \multicolumn{1}{c|}{126\degree} & \multicolumn{1}{c|}{144\degree} & \multicolumn{1}{c|}{162\degree} & 180\degree &  \\ \lhline
\multicolumn{1}{c|}{\multirow{5}{*}{\rotatebox[origin=c]{0}{NM\#5-6}}} & GaitSet~\cite{gaitset_ChaoHZF19} & \multicolumn{1}{c|}{90.8} & \multicolumn{1}{c|}{97.9} & \multicolumn{1}{c|}{\best{99.4}} & \multicolumn{1}{c|}{96.9} & \multicolumn{1}{c|}{\cc 93.6} & \multicolumn{1}{c|}{\cc 91.7} & \multicolumn{1}{c|}{\cc 95.0} & \multicolumn{1}{c|}{97.8} & \multicolumn{1}{c|}{98.9} & \multicolumn{1}{c|}{96.8} & 85.8 & 95.0 \\ 
\multicolumn{1}{c|}{}  & GaitPart~\cite{gaitpart_FanPC0HCHLH20_cvpr}  & \multicolumn{1}{c|}{94.1} & \multicolumn{1}{c|}{98.6} & \multicolumn{1}{c|}{\better{99.3}} & \multicolumn{1}{c|}{\best{98.5}} & \multicolumn{1}{c|}{\cc 94.0} & \multicolumn{1}{c|}{\cc 92.3} & \multicolumn{1}{c|}{\cc 95.9} & \multicolumn{1}{c|}{98.4} & \multicolumn{1}{c|}{99.2} & \multicolumn{1}{c|}{97.8} & 90.4 & 96.2 \\ 
\multicolumn{1}{c|}{} & GaitEdge$^{\diamond}$~\cite{gaitedge_eccv22_liang} & \multicolumn{1}{c|}{\best{97.2}} & \multicolumn{1}{c|}{\best{99.1}} & \multicolumn{1}{c|}{99.2} & \multicolumn{1}{c|}{\better{\cc 98.3}} & \multicolumn{1}{c|}{\best{\cc 97.3}} & \multicolumn{1}{c|}{\better{\cc 95.5}} & \multicolumn{1}{c|}{\better{\cc 97.1}} & \multicolumn{1}{c|}{\best{\cc 99.4}} & \multicolumn{1}{c|}{\best{\cc 99.3}} & \multicolumn{1}{c|}{98.5} & \best{96.4} & \better{97.9} \\ 
\multicolumn{1}{c|}{} & GaitGL~\cite{gl_LinZ021_iccv} & \multicolumn{1}{c|}{96.0} & \multicolumn{1}{c|}{98.3} & \multicolumn{1}{c|}{99.0} & \multicolumn{1}{c|}{97.9} & \multicolumn{1}{c|}{\cc 96.9} & \multicolumn{1}{c|}{\cc 95.4} & \multicolumn{1}{c|}{\cc 97.0} & \multicolumn{1}{c|}{98.9} & \multicolumn{1}{c|}{\best{99.3}} & \multicolumn{1}{c|}{\better{98.8}} & 94.0 & 97.4 \\ 
\multicolumn{1}{c|}{} & \textbf{\methodname} & \multicolumn{1}{c|}{\better{96.3}} & \multicolumn{1}{c|}{\better{98.8}} & \multicolumn{1}{c|}{99.1} & \multicolumn{1}{c|}{98.1} & \multicolumn{1}{c|}{\better{\cc 97.2}} & \multicolumn{1}{c|}{\best{\cc 96.5}} & \multicolumn{1}{c|}{\best{\cc 98.2}} & \multicolumn{1}{c|}{\better{99.1}} & \multicolumn{1}{c|}{\best{99.3}} & \multicolumn{1}{c|}{\best{99.2}} & \better{95.9} & \best{98.0} \\ \hline
\multicolumn{1}{c|}{\multirow{5}{*}{\rotatebox[origin=c]{0}{BG\#1-2}}} & GaitSet~\cite{gaitset_ChaoHZF19} & \multicolumn{1}{c|}{83.8} & \multicolumn{1}{c|}{91.2} & \multicolumn{1}{c|}{91.8} & \multicolumn{1}{c|}{\cc 88.8} & \multicolumn{1}{c|}{\cc 83.3} & \multicolumn{1}{c|}{\cc 81.0} & \multicolumn{1}{c|}{\cc 84.1} & \multicolumn{1}{c|}{\cc 90.0} & \multicolumn{1}{c|}{92.2} & \multicolumn{1}{c|}{94.4} & 79.0 & 87.2 \\ 
\multicolumn{1}{c|}{} & GaitPart~\cite{gaitpart_FanPC0HCHLH20_cvpr} & \multicolumn{1}{c|}{89.1} & \multicolumn{1}{c|}{94.8} & \multicolumn{1}{c|}{96.7} & \multicolumn{1}{c|}{\cc 95.1} & \multicolumn{1}{c|}{\cc 88.3} & \multicolumn{1}{c|}{\cc 84.9} & \multicolumn{1}{c|}{\cc 89.0} & \multicolumn{1}{c|}{\cc 93.5} & \multicolumn{1}{c|}{96.1} & \multicolumn{1}{c|}{93.8} & 85.8 & 91.5 \\ 
\multicolumn{1}{c|}{} & GaitEdge$^{\diamond}$~\cite{gaitedge_eccv22_liang} & \multicolumn{1}{c|}{\best{95.3}} & \multicolumn{1}{c|}{\best{97.4}} & \multicolumn{1}{c|}{\best{98.4}} & \multicolumn{1}{c|}{\best{97.6}} & \multicolumn{1}{c|}{\better{\cc 94.3}} & \multicolumn{1}{c|}{\better{\cc 90.6}} & \multicolumn{1}{c|}{\better{\cc 93.1}} & \multicolumn{1}{c|}{\best{\cc 97.8}} & \multicolumn{1}{c|}{\best{99.1}} & \multicolumn{1}{c|}{\best{98.0}} & \best{95.0} & \best{96.1}\\
\multicolumn{1}{c|}{} & GaitGL~\cite{gl_LinZ021_iccv} & \multicolumn{1}{c|}{92.6} & \multicolumn{1}{c|}{\better{96.6}} & \multicolumn{1}{c|}{96.8} & \multicolumn{1}{c|}{\cc 95.5} & \multicolumn{1}{c|}{\cc 93.5} & \multicolumn{1}{c|}{\cc 89.3} & \multicolumn{1}{c|}{\cc 92.2} & \multicolumn{1}{c|}{\cc 96.5} & \multicolumn{1}{c|}{98.2} & \multicolumn{1}{c|}{96.9} & 91.5 & 94.5 \\ 
\multicolumn{1}{c|}{} & \textbf{\methodname} & \multicolumn{1}{c|}{\better{93.7}} & \multicolumn{1}{c|}{       96.4} & \multicolumn{1}{c|}{\better{97.4}} & \multicolumn{1}{c|}{\better{\cc 97.2}} & \multicolumn{1}{c|}{\best{\cc 96.2}} & \multicolumn{1}{c|}{\best{\cc 93.4}} & \multicolumn{1}{c|}{\best{\cc 95.5}} & \multicolumn{1}{c|}{\best{\cc 97.8}} & \multicolumn{1}{c|}{\better{98.4}} & \multicolumn{1}{c|}{\better{97.8}} & \better{93.1} & \best{96.1} \\ \hline
\multicolumn{1}{c|}{\multirow{5}{*}{\rotatebox[origin=c]{0}{CL\#1-2}}} & GaitSet\cite{gaitset_ChaoHZF19} & \multicolumn{1}{c|}{61.4} & \multicolumn{1}{c|}{\cc 75.4} & \multicolumn{1}{c|}{\cc 80.7} & \multicolumn{1}{c|}{\cc 77.3} & \multicolumn{1}{c|}{\cc 72.1} & \multicolumn{1}{c|}{\cc 70.1} & \multicolumn{1}{c|}{\cc 71.5} & \multicolumn{1}{c|}{\cc 73.5} & \multicolumn{1}{c|}{\cc 73.5} & \multicolumn{1}{c|}{\cc 68.4} & \cc 50.0 & 70.4 \\ 
\multicolumn{1}{c|}{} & GaitPart~\cite{gaitpart_FanPC0HCHLH20_cvpr} & \multicolumn{1}{c|}{70.7} & \multicolumn{1}{c|}{\cc 85.5} & \multicolumn{1}{c|}{\cc 86.9} & \multicolumn{1}{c|}{\cc 83.3} & \multicolumn{1}{c|}{\cc 77.1} & \multicolumn{1}{c|}{\cc 72.5} & \multicolumn{1}{c|}{\cc 76.9} & \multicolumn{1}{c|}{\cc 82.2} & \multicolumn{1}{c|}{\cc 83.8} & \multicolumn{1}{c|}{\cc 80.2} & \cc 66.5 & 78.7 \\ 
\multicolumn{1}{c|}{} & GaitEdge$^{\diamond}$~\cite{gaitedge_eccv22_liang} & \multicolumn{1}{c|}{\best{84.3}} & \multicolumn{1}{c|}{\better{92.8}} & \multicolumn{1}{c|}{\better{94.3}} & \multicolumn{1}{c|}{\cc \better{92.2}} & \multicolumn{1}{c|}{\cc \better{84.6}} & \multicolumn{1}{c|}{\cc \better{83.0}} & \multicolumn{1}{c|}{\cc 83.0} & \multicolumn{1}{c|}{\cc \better{87.5}} & \multicolumn{1}{c|}{\cc \better{87.4}} & \multicolumn{1}{c|}{\better{85.9}} & \better{75.0} & \better{86.4} \\
\multicolumn{1}{c|}{} & GaitGL~\cite{gl_LinZ021_iccv} & \multicolumn{1}{c|}{76.6} & \multicolumn{1}{c|}{\cc 90.0} & \multicolumn{1}{c|}{\cc 90.3} & \multicolumn{1}{c|}{\cc 87.1} & \multicolumn{1}{c|}{\cc 84.5} & \multicolumn{1}{c|}{\cc 79.0} & \multicolumn{1}{c|}{\cc \better{84.1}} & \multicolumn{1}{c|}{\cc 87.0} & \multicolumn{1}{c|}{\cc 87.3} & \multicolumn{1}{c|}{\cc 84.4} & \cc 69.5 & 83.6 \\ 
\multicolumn{1}{c|}{} & \textbf{\methodname} & \multicolumn{1}{c|}{\better{79.6}} & \multicolumn{1}{c|}{\best{\cc 93.4}} & \multicolumn{1}{c|}{\best{\cc 95.0}} & \multicolumn{1}{c|}{\best{\cc 92.4}} & \multicolumn{1}{c|}{\best{\cc 88.4}} & \multicolumn{1}{c|}{\best{\cc 82.5}} & \multicolumn{1}{c|}{\best{\cc 86.9}} & \multicolumn{1}{c|}{\best{\cc 91.4}} & \multicolumn{1}{c|}{\best{\cc 93.9}} & \multicolumn{1}{c|}{\best{\cc 90.1}} &  \best{75.3} & \best{88.1} \\\shline
\end{tabular}
}
\caption{Averaged rank-1 accuracies on CASIA-B under three different conditions, excluding identical-view cases. The superscript $^{\diamond}$ notes that the input modality is RGB.} 
\label{table: casia-b}
\end{table*}

\begin{table*}[h]
\centering
\resizebox{0.82\textwidth}{!}{
\begin{tabular}{c|cccccccccccccc|c}
\shline
\multirow{2}{*}{Method} & \multicolumn{14}{c|}{Probe view} & \multirow{2}{*}{Mean} \\ \cline{2-15}
 & \multicolumn{1}{c|}{0\degree} & \multicolumn{1}{c|}{15\degree} & \multicolumn{1}{c|}{30\degree} & \multicolumn{1}{c|}{45\degree} & \multicolumn{1}{c|}{60\degree} & \multicolumn{1}{c|}{75\degree} & \multicolumn{1}{c|}{90\degree} & \multicolumn{1}{c|}{180\degree} & \multicolumn{1}{c|}{195\degree} & \multicolumn{1}{c|}{210\degree} & \multicolumn{1}{c|}{225\degree} & \multicolumn{1}{c|}{240\degree} & \multicolumn{1}{c|}{255\degree} & 270\degree &  \\ \hline
GaitSet$^\star$~\cite{gaitset_ChaoHZF19}  & \multicolumn{1}{c|}{78.7} & \multicolumn{1}{c|}{87.4} & \multicolumn{1}{c|}{89.8} & \multicolumn{1}{c|}{90.0} & \multicolumn{1}{c|}{87.8} & \multicolumn{1}{c|}{88.5} & \multicolumn{1}{c|}{97.5} & \multicolumn{1}{c|}{81.3} & \multicolumn{1}{c|}{86.2} & \multicolumn{1}{c|}{88.9} & \multicolumn{1}{c|}{89.1} & \multicolumn{1}{c|}{87.1} & \multicolumn{1}{c|}{87.6} & 86.1 & 86.9 \\ \hline
GaitPart$^\star$~\cite{gaitpart_FanPC0HCHLH20_cvpr}& \multicolumn{1}{c|}{82.1} & \multicolumn{1}{c|}{88.8} & \multicolumn{1}{c|}{90.7} & \multicolumn{1}{c|}{90.8} & \multicolumn{1}{c|}{89.5} & \multicolumn{1}{c|}{89.7} & \multicolumn{1}{c|}{89.1} & \multicolumn{1}{c|}{84.7} & \multicolumn{1}{c|}{87.4} & \multicolumn{1}{c|}{89.9} & \multicolumn{1}{c|}{90.0} & \multicolumn{1}{c|}{88.7} & \multicolumn{1}{c|}{88.9} & 87.8 & 88.4 \\ \hline
GaitGL$^\star$~\cite{gl_LinZ021_iccv} & \multicolumn{1}{c|}{\better{84.3}} & \multicolumn{1}{c|}{\better{89.9}} & \multicolumn{1}{c|}{\better{91.1}} & \multicolumn{1}{c|}{\better{91.4}} & \multicolumn{1}{c|}{\better{90.9}} & \multicolumn{1}{c|}{\better{90.6}} & \multicolumn{1}{c|}{\better{90.2}} & \multicolumn{1}{c|}{\better{88.3}} & \multicolumn{1}{c|}{\better{88.1}} & \multicolumn{1}{c|}{\better{90.3}} & \multicolumn{1}{c|}{\better{90.4}} & \multicolumn{1}{c|}{\better{89.6}} & \multicolumn{1}{c|}{\better{89.4}} & \better{88.6} & \better{89.5} \\ \hline

\textbf{\methodname} & \multicolumn{1}{c|}{\best{87.1}} & \multicolumn{1}{c|}{\best{91.0}} & \multicolumn{1}{c|}{\best{91.4}} & \multicolumn{1}{c|}{\best{91.8}} & \multicolumn{1}{c|}{\best{91.7}} & \multicolumn{1}{c|}{\best{91.3}} & \multicolumn{1}{c|}{\best{91.1}} & \multicolumn{1}{c|}{\best{90.3}} & \multicolumn{1}{c|}{\best{89.6}} & \multicolumn{1}{c|}{\best{90.7}} & \multicolumn{1}{c|}{\best{90.8}} & \multicolumn{1}{c|}{\best{90.5}} & \multicolumn{1}{c|}{\best{90.2}} & {\best{89.9}} & {\best{90.5}} \\ \hline
\shline
\end{tabular}
}
\caption{Averaged rank-1 accuracies on OU-MVLP, excluding identical-view cases. The superscript $^\star$ notes the average rank-1 accuracies are reproduced results in our test sets for a fair comparison.}
\label{table: oumvlp}
\end{table*}

\subsection{Implementation Details}
The gait silhouettes are normalized before being fed to the network with a fixed input size, $64\times44$. And the batch size $(p,k)$ is $(8,8)$ in CASIA-B dataset and $(32,8)$ in OU-MVLP dataset, respectively. The optimizer is Adam and the learning rate is $1e$-$4$ in all experiments. For experiments on CASIA-B, the training iterations is set to $80$k and the learning rate decay to $1e$-$5$ after $70$k. For the OU-MVLP, the total iterations are $90$k, and the learning rate decay to $1e$-$5$ after $80$k. In our \methodname network for CASIA-B, the number of output channel are $32,128,256,256$ for each stage respectively. Since the OU-MVLP is 20 times bigger than CASIA-B, we directly double the convolution layers in each block. 
Thus the output channel of each stage holds $32,128, 256, 256$. After the first stage, a channel interaction layer implemented by a 1$\times$1 convolution is added on OU-MVLP. Other hyperparameters are following the backbone's settings. We use four NVIDIA GeForce RTX 3090 GPUs for training \methodname. 

\subsection{Comparison with the State-of-the-art Method}
From Table~\ref{table: casia-b}, it can be seen that the average rank-1 accuracies of \methodname outperforms GaitGL by 0.6$\%$, 1.6$\%$ and 4.5$\%$ in the NM, BG, and CL conditions and implies the superiority of \methodname. 
Noteworthy, the performance on $90\degree$ has been upgraded by our \methodname, exceeding GaitGL by 1.1$\%$~(NM), 4.1$\%$~(BG) and 3.5$\%$~(CL) since the evident legging movement in this viewpoint benefits \methodname.
Moreover, although GaitEdge takes the more informative RGB image as input, our \methodname with the silhouette as input still surpasses it, achieving $1.7\%$ higher in the CL. 
The superior performance indicates that the \methodname has a strong representation ability, even under challenging conditions. 

The experimental results on OU-MVLP~(Table~\ref{table: oumvlp}) show great progress than GaitGL by $1.0\%$, demonstrating the superiority of \methodname. Note that when discarding the illegal sequences, the average rank-1 accuracy will rise to $97.1\%$.

\subsection{Ablation Studies}

\begin{table}[]
\centering
\resizebox{0.3\textwidth}{!}{
\begin{tabular}{lcccc}
\shline
\multicolumn{1}{l|}{Methods} & \multicolumn{1}{c|}{NM} & \multicolumn{1}{c|}{BG} & \multicolumn{1}{c|}{CL} & Mean \\ \lhline
\multicolumn{1}{l|}{GaitGL~\cite{gl_LinZ021_iccv}} & \multicolumn{1}{c|}{97.4} & \multicolumn{1}{c|}{94.5} & \multicolumn{1}{c|}{83.6} & 91.8 \\ \hline
\multicolumn{1}{l|}{\quad\quad w/  LAGM~\cite{lagana_cvpr22_chai}} & \multicolumn{1}{c|}{96.8} & \multicolumn{1}{c|}{93.1} & \multicolumn{1}{c|}{84.7} & 91.5 \\ 
\multicolumn{1}{l|}{\quad\quad w/ \dynamicmodule} & \multicolumn{1}{c|}{\best{98.1}} & \multicolumn{1}{c|}{95.8} & \multicolumn{1}{c|}{87.0} & 93.6 \\ 
\multicolumn{1}{l|}{\quad\quad w/ \enhancemodule} & \multicolumn{1}{c|}{97.2} & \multicolumn{1}{c|}{94.7} & \multicolumn{1}{c|}{84.3} & 92.1 \\ 
\multicolumn{1}{l|}{\quad\quad \empmethodname} & \multicolumn{1}{c|}{98.0} & \multicolumn{1}{c|}{\best{96.1}} & \multicolumn{1}{c|}{\best{88.1}} & \best{94.1} \\ \lhline
\multicolumn{1}{l|}{GaitSet~\cite{gaitset_ChaoHZF19}} & \multicolumn{1}{c|}{95.0} & \multicolumn{1}{c|}{87.2} & \multicolumn{1}{c|}{70.4} & 84.2 \\ \hline 
\multicolumn{1}{l|}{\quad\quad w/  LAGM~\cite{lagana_cvpr22_chai}} & \multicolumn{1}{c|}{95.0} & \multicolumn{1}{c|}{87.3} & \multicolumn{1}{c|}{\best{73.3}} & 85.2 \\ 
\multicolumn{1}{l|}{\quad\quad GaitSet-\textbf{M}} & \multicolumn{1}{c|}{\best{95.6}} & \multicolumn{1}{c|}{\best{89.0}} & \multicolumn{1}{c|}{73.0} & \best{85.9} \\ 
 \shline 
\end{tabular}
}
\caption{The top half of the table ablates the effect of \dynamicmodule and \enhancemodule. The bottom half shows the results of applying the proposed motion modeling to GaitSet.}
\vspace{-16pt}
\label{table: network_ablation}
\end{table}

To explore the contributions of \dynamicmodule and \enhancemodule, we design the ablation studies presented in Table~\ref{table: network_ablation}. It is worth noting that the proposed \methodname outperforms GaitGL, even only leaving one motion-aware module. 

By comparing the designed \dynamicmodule and \enhancemodule, we observe that the SiMo contributes more than the FeMo. The possible reason is that the motion information is missed by the smoothed effect of convolution as the network deepens. Therefore, although it boosts the motion-related element, the features themselves contain little motion information. 

To further verify the application to other backbones, we apply our \dynamicmodule and \enhancemodule to GaitSet, shown on the bottom part of Table~\ref{table: network_ablation}. It can be found that the averaged rank-1 accuracy has been upgraded by $1.7\%$ with our modules. By comparing with LAGM~\cite{lagana_cvpr22_chai}, which regards similarity as motion information, our method based on exploring the pattern of the temporal changes can achieve more performance improvement.

\vspace{-10pt}
\section{Conclusion}

This paper proposed a novel motion modeling to enjoy the discrimination and robustness of the motion information. Specifically, one silhouette-level motion extractor and feature-level motion enhancement module have been devised to facilitate the motion features in the whole feature extraction stages. Extensive experimental results verify that motion matters in gait recognition and demonstrate the superiority of our motion modeling, which may serve as a plug-and-play module in future model designs.

\vfill\pagebreak

\bibliographystyle{IEEEbib}
\bibliography{refs}

\end{document}